\documentclass[runningheads]{llncs}
\usepackage{graphicx}
\newcommand\tab[1][.5cm]{\hspace*{#1}}
\usepackage{amsmath}
\usepackage{graphicx} 
\usepackage[colorinlistoftodos]{todonotes}
\usepackage{algorithm}
\usepackage[noend]{algpseudocode}
\usetikzlibrary{positioning}
\usepackage{siunitx}
\usepackage[T1]{fontenc}
\usepackage{amsfonts}
\usepackage{tabularx}
\usepackage{ragged2e}
\usepackage{subfig}
\usepackage{tikz}
\usetikzlibrary{shapes,arrows,positioning,calc}
\usepackage{wrapfig}

\begin{document}
\title{Deep Eikonal Solvers}
%
%
%
\author{Moshe Lichtenstein \and Gautam Pai \and Ron Kimmel}
\authorrunning{Lichtenstein et al.}
 \institute{Technion - Israel Institute of Technology, Haifa, Israel
\\ 
 \email{\{smosesli, paigautam, ron\}@cs.technion.ac.il}
}
%
\maketitle              
\begin{abstract}
A deep learning approach to numerically approximate the solution to the Eikonal equation is introduced. 
The proposed method is built on the fast marching scheme which comprises of two components: a local numerical solver and an update scheme. 
We replace the formulaic local numerical solver with a trained neural network to provide highly accurate estimates of local distances for a variety of different geometries and sampling conditions. 
Our learning approach generalizes not only to flat Euclidean domains but also to curved surfaces enabled by the incorporation of certain invariant features in the neural network architecture. 
We show a considerable gain in performance, validated by smaller errors and higher orders of accuracy for the numerical solutions of the Eikonal equation computed on different surfaces
The proposed approach leverages the approximation power of neural networks to enhance the performance of numerical algorithms, thereby, connecting the somewhat disparate themes of numerical geometry and learning.

\end{abstract}
\section{Introduction}
Fast and accurate computation of distances is fundamental to innumerable problems in computer science. 
Specifically, in the sub-fields of computer vision, geometry processing and robotics, distance computation is imperative for applications like navigating robots \cite{lee2016structured,kimmel1998multivalued,kimmel2001optimal}, video object segmentation \cite{wang2015saliency}, image segmentation \cite{chen2017fast} and shape matching \cite{younes2012spaces}. 
A distance function has a gradient with unit magnitude at every point in the domain, therefore, they are computed by estimating the viscosity solutions to a Hamilton-Jacobi type non-linear PDE called the Eikonal equation.
Fast marching methods are the prominent numerical schemes to estimate distance functions on discretized Euclidean domains as well as triangulated curved surfaces. 
Fast marching methods are known to have a $\mathcal{O}(N \log N)$ computational complexity (for a discrete domain consist in $N$ sample points). 
Different versions of the fast marching algorithm have been developed yielding different accuracies $\mathcal{O}(h)$ \cite{sethian1996fast}, $\mathcal{O}(h^2)$ \cite{sethian1999level} and $\mathcal{O}(h^3)$ \cite{ahmed2011third} for 2D Cartesian grids and $\mathcal{O}(h)$ for triangulated surfaces \cite{kimmel1998computing}, where $h$ is the resolution of the discretization.
A fast marching method comprises of a local numerical solver and an update step. 
The local numerical solver approximates the gradient locally to estimate the distance of a point in the advancing unit speed wavefront. 
Similar to the strategy employed in the celebrated Dijkstra's algorithm \cite{dijkstra1959note}, the update step involves selecting the least distant point in order to propagate the wavefront. 
The accuracy of fast marching scheme relies heavily on the local solver which is responsible for approximations to the gradient at the point of the advancing front \cite{tsitsiklis1995efficient,sethian1996fast}.

The success of deep learning has shown that neural networks can be utilized as powerful function approximators of complex attributes governing various visual and auditory phenomena. 
The availability of large amounts of data and computational power, coupled with parallel streaming architectures and improved optimization techniques, have led to computational frameworks that efficiently exploit their representational power. 
However, the utility of neural networks to provide accurate solutions to complex PDE's is still a very nascent topic in computational research. 
Notable efforts include methods like \cite{sirignano2018dgm} which attempt to solve high-dimensional parabolic partial differential equations, whereas \cite{long2017pde} demonstrates the learning of differential operators from data as convolution kernels.

The main contribution of this paper is to develop a deep-learning infrastructure that enables accurate distance computation. 
Using the identical computational schematic of the fast marching method, the basic premise of this paper is that one can \emph{learn} the local numerical solver by training a neural network and use it in conjunction with the upwind update scheme. 
Experiments in sections \ref{sect:cartesian} and \ref{sect:results} demonstrate lower errors and higher order of accuracy in the solutions. 
Importantly, we show that our method generalizes to different geometries and sampling conditions. 
\section{Background}
\subsection{The Eikonal Equation}
Given a domain $\Omega \in \mathbb{R}^n$ and a curve or surface $\Gamma \subset \Omega$, we wish to find the distance, denoted by $u(x)$, for each $x \in \Omega$ from the given $\Gamma$. 
The distance function satisfies 
\begin{eqnarray}
\label{eq:eq1}
 |\nabla u(x)| &=& 1 , \,\,\, \,\,\,\, x \in \Omega \setminus \Gamma \cr
 u(x) &=& 0 , \,\,\, \,\,\,\, x \in \Gamma,
 \end{eqnarray} 
where $\nabla$ refers to the gradient operator. 
Equation (\ref{eq:eq1}) can be generalized to arbitrary boundary conditions $u(x)=g(x), \,\, x \in \Gamma$, and also to weighted domains where the local metric at a point $x$ is defined by a positive scalar function $F:\Omega \rightarrow \mathbb{R}^+$. 
As shown in \cite{crandall1983viscosity,ishii1984uniqueness,ishii1987simple}, the viscosity solution for Equation (\ref{eq:eq1}) is unique.

Next, we consider curved manifolds, specifically, surfaces embedded in $\mathbb{R}^3$. 
Let ${\cal M}$ be a Riemannian manifold with metric tensor $G$.  
The geodesic distance function $u: \mathcal{M} \rightarrow \mathbb{R}^+$ satisfies
\begin{eqnarray}
\label{eq:eikonal_mesh}
|\nabla_G u(x)| &=& 1 , \,\,\,\, x \in \mathcal{M}  \setminus \Gamma \cr
 u(x) &=& 0 , \,\,\,\, x \in \Gamma,
 \end{eqnarray}
where $\nabla_G$ represents the gradient operator defined on the Riemannian manifold. 

\subsection{Numerical Approximation}
Consider the task of numerically approximating the function $u(x)$ in Equation (\ref{eq:eq1}). 
Let $u_{i,j} \equiv u(ih, jh)$, where $h$ defines the distance between neighboring grid points along the $x$ and $y$ axis. 
The following numerical approximation is shown \cite{osher1988fronts,rouy1992viscosity} to pick the unique viscosity solution of Equation (\ref{eq:eq1}),
\begin{equation}
\label{eq:approx}
|\nabla u_{i,j}|^2 \approx 
	\max(D^{-x}_{i,j}u, -D^{+x}_{i,j}u, 0)^2 +
	\max(D^{-y}_{i,j}u, -D^{+y}_{i,j}u, 0)^2.
\end{equation}
Where $\{D^{+x}_{i,j},D^{-x}_{i,j}\}, \{D^{+y}_{i,j},D^{-y}_{i,j}\}$ are the forward and backward difference operators for point $(i,j)$ along the $x$ and $y$ directions, respectively. 
Based on approximation (\ref{eq:approx}), the solution of $u_{i,j}$, given the $u$ values at its four neighbors $\{u_{i+1,j}, u_{i-1,j}, u_{i,j+1}, u_{i,j-1}\}$ requires a solution of the quadratic equation  
 \begin{eqnarray}
\label{eq:eqXX}
	\max(D^{-x}_{i,j}u, -D^{+x}_{i,j}u, 0)^2 +
	\max(D^{-y}_{i,j}u, -D^{+y}_{i,j}u, 0)^2 &=& 1.
\end{eqnarray}

\begin{table}[t]
\vspace{-0.56cm}
\centering
\includegraphics[scale = 0.35]{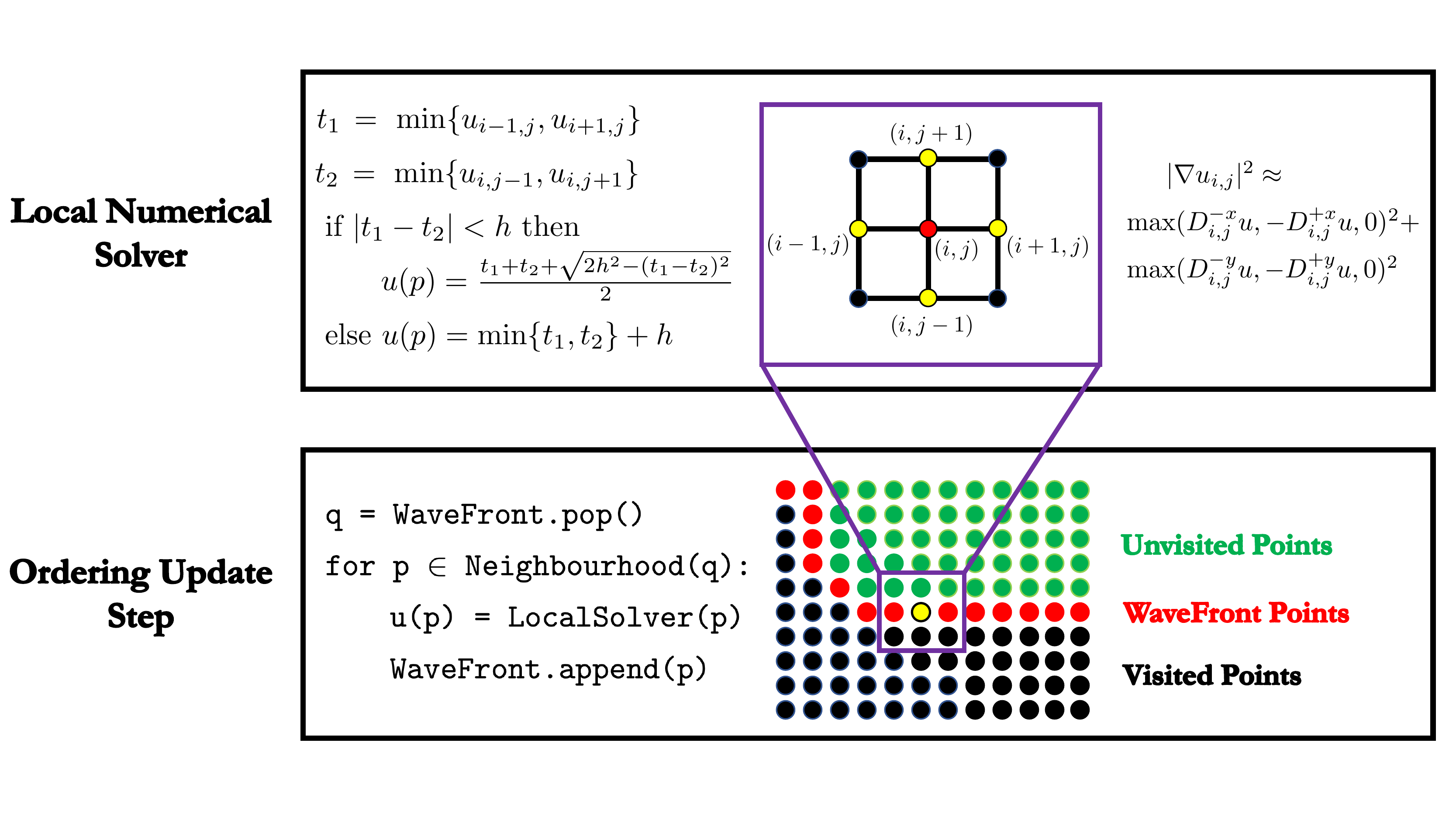}
\vspace{-0.5cm}
\captionof{figure}{{\bf Schematic for Fast Marching :} The numerical solver approximates the gradient locally while the ordering scheme advances the unit speed wavefront. 
The suggested method employs a trained neural network as a Local Solver.}
\label{fig:fmm_schematic}
\vspace{-.7cm}
\begin{algorithm}[H] 
\captionof{algorithm}{Eikonal Estimation on Discretized Domain }
\label{algo:dgc}
	\begin{algorithmic}[1]   
		\State \textbf{Initialize:}
        \State $u(p)=0$, \, Tag $p$ as $visited$; $\forall p  \in \mathfrak{s}$
		\State $u(p)=\infty$, Tag $p$ as $unvisited$; $\forall p \in \mathcal{S} \setminus \mathfrak{s}$
		\While{there is a non-$visited$ point}
        \newline
        \tab Denote the points adjacent to the newly $visited$ points as $\mathcal{A}$
        \ForAll {$p \in \mathcal A $}
        \State  Estimate $u(p)$ based on $visited$ points.\label{step6}
		\State Tag $p$ as $wavefront$
        \EndFor
		\State Tag the least distant $wavefront$ point as $visited$.\label{step7}
		\EndWhile
        \State \Return $u$
	\end{algorithmic}
\end{algorithm} 
\vspace{-1.2cm}
\end{table}
Typically, the approximation in Equation (\ref{eq:approx}) is based on first order Taylor series expansion of $u_{i,j}$. However, more sophisticated approximations such as those using two points \cite{sethian1999level} or three points \cite{ahmed2011third} away from $(i,j)$ in each direction, or using diagonal neighbors \cite{hassouna2007multistencils} yield more accurate solutions. Unlike Euclidean domains which enjoy the benefit of regular sampling, developing local approximations similar to Equation (\ref{eq:approx}) for curved manifolds are evidently more challenging due to the lack of a universal regular sampling strategy.

\subsection{Fast Eikonal Solvers}
Fast Eikonal solvers estimate distance functions in linear or quasilinear time. They typically comprise of two computational parts: a local numerical solver and an ordering/update method. The local solver provides an approximation for the distance function at a desired point and the ordering method chooses the next point to approximate. 
Fast marching methods \cite{tsitsiklis1995efficient,sethian1996fast} is the most prominent subclass of fast Eikonal solvers that use approximations similar to Equation (\ref{eq:approx}) for computing the distance function at a point using the known distances of its neighbors. 
By employing Dijkstra's ordering strategy, fast marching schemes display an \emph{upwind} nature, where the solution can be seen to grow outward from the source, equivalent to the propagation of a unit speed wavefront (See Figure \ref{fig:fmm_schematic}).
The other prominent subclass of fast Eikonal solvers are the fast sweeping methods \cite{zhao2000implicit,zhao2005fast,kimmel2006method,weber2008parallel},  
iterative schemes that use alternating sweeping ordering. Some variants \cite{li2008second} of fast sweeping establish their local solver on finite element techniques, however most of them use the same upwind difference local solver as fast marching.

Kimmel and Sethian \cite{kimmel1998computing} extended the fast marching scheme to approximate geodesic distances on triangulated surfaces by treating the sampled manifold as piecewise planar. In order to estimate $u(p)$ for some mesh point $p$, each triangle involving $p$ is processed independently. Given a triangle comprising of the points $p_1,p_2,p_3$, the distance $u(p_1)$ is estimated based on $u(p_2), u(p_3)$ by approximating Equation (\ref{eq:eq1}) on the triangle's plane. 
The minimum of the solutions computed on all the triangles involving $p$ is chosen as the final distance assigned to $p$.  
This algorithm is analogous to the first-order fast marching method that operates on regular grids and therefore has an accuracy $\mathcal{O}(h)$.
\section{Deep Eikonal Solver}

We present a fast Eikonal solver estimating the geodesic distance for a discrete set of points $\mathcal{S}$ from a subset of source points $\mathfrak{s} \subset \mathcal{S}$. 
The local solver (step \ref{step6} in Algorithm \ref{algo:dgc}) in our scheme is represented by a neural network. The process of choosing subsequent points for evaluation is similar to Dijkstra's algorithm \cite{dijkstra1959note}, according to which each point $p \in \mathcal{S}$ is tagged into one of three disjoint sets, 
\begin{enumerate}
\item {\bf Visited:} where $u(p)$ is already computed (Black points in Figure \ref{fig:fmm_schematic}).
\item {\bf WaveFront:} where $u(p)$ calculation is in process. (Red points in Figure \ref{fig:fmm_schematic}). 
\item {\bf Unvisited:} where $u(p)$ is yet to be determined (Green points in Figure \ref{fig:fmm_schematic}).  
\end{enumerate}
Since we use the same update method used in the fast marching scheme, the computational complexity of our method is $\mathcal{O}(N \log N)$. 
In Section \ref{sect:train} we describe the general philosophy behind training the network. In Sections \ref{sect:cartesian} and \ref{sect:results} we elaborate on the exact procedure and architecture of the network specific to the corresponding domain. 

\subsection{Training the Local Solver}
\label{sect:train}
Our basic premise is that the local solver can be \emph{learned} using a neural network. The input to the local solver which is in action at a certain point $p \in {\mbox{WaveFront}}$ is the set of points in its neighborhood denoted by $\mathcal{N}(p) = \{p_1, p_2, \ldots, p_M\}$ and their corresponding distances $\{u(p_1), u(p_2), \ldots , u(p_M)\}$. 
Based on the local geometry of the $\mbox{Visited}$ points in $p$'s neighborhood, an estimate of the local distance is outputted by the local solver. See Figures \ref{fig:fmm_schematic}, \ref{fig:cartesian} and \ref{fig:mesh} for a visual schematic.

Following a supervised training procedure, we train the local solver with examples containing the \emph{ground truth} distances. By \emph{ground truth} we refer to the most accurate available solution for distance (elaborated further in sections \ref{sect:cartesian} and \ref{sect:mesh}). For a given point $p$, denote this distance as $u_{gt}(p)$. The network inputs neighborhood information: 
$\{p_1, u_{gt}(p_1), \ldots, p_M, u_{gt}(p_M)\}$ and is trained to minimize the difference between its output and the corresponding ground truth distance at the point $p : u_{gt} (p)$. 

To simulate the propagating unit speed wavefront in various representative scenarios, we employ the following strategy. We develop a variety of different sources and construct a dataset of local neighborhood patches at different locations relative to the sources. We allow a patch point to participate in the prediction of $u(p)$ according to the following rule. 
\begin{equation}
 q \in {\mbox{Visited}} \;\; \text{if} \;\; u_{gt}(q) < u_{gt}(p).
\label{eq:gt_rule}
\end{equation}
Therefore, given a point $p$ and a patch of points constituting its neighborhood: $\{p_1, p_2, \ldots ,p_M\}$,  the network inputs $\{p_1,u_{gt}(p_1), \ldots,p_M, u_{gt}(p_M)\}$, where all points that are declared ${\mbox{Visited}}$ according to rule \ref{eq:gt_rule} 
input their respective ground truth distance. 
For points $q \in \mathcal{N}(p)$ that do not satisfy condition \ref{eq:gt_rule}, we simulate that the wavefront starting from the source arrives at their location later than it arrives to $p$. 
Hence, such points are not considered ${\mbox{Visited}}$ and they input a constant value to the network.
\begin{equation}
\text{If } q \notin {\mbox{Visited}} ; \;\; u_{gt}(q) = C
\label{eq:gt_rule_2}
\end{equation}
By employing the rules \ref{eq:gt_rule} and \ref{eq:gt_rule_2} stated above, we create a rich database of such local patches coupled with the desired outputs. The parameters of the network $\Theta$ are optimized to minimize the MSE loss
\begin{equation}
L(\Theta) = \underset{i}{\sum} \Big( f_{\Theta} \big( p_{i1}, u_{gt}(p_{i1}), \ldots,p_{iM}, u_{gt}(p_{iM}) \big) - u_{gt}(p_i) \Big)^2,
\end{equation}
over a dataset comprising of a variety of points $p_i$ and their corresponding neighborhoods. 

The intuition behind our approach is twofold. 
First, we expect that using a larger local support leads to a more accurate estimate of the solution to the differential equation. 
Secondly, by training the network to follow a \emph{ground truth} distance, we avoid the need to develop a sophisticated formula for locally approximating the gradient with high accuracy. 
Rather, we expect the network to \emph{learn} the necessary relationship between the local patch and the ground truth distance. 
As demonstrated in our results section, our learning-based approach generalizes to different scenarios including different shapes and different datasets. 

\begin{table}[t]
\vspace{-0.55cm}
\centering
\subfloat[]{\hspace{-2.5cm}{\includegraphics[width=6.5cm]{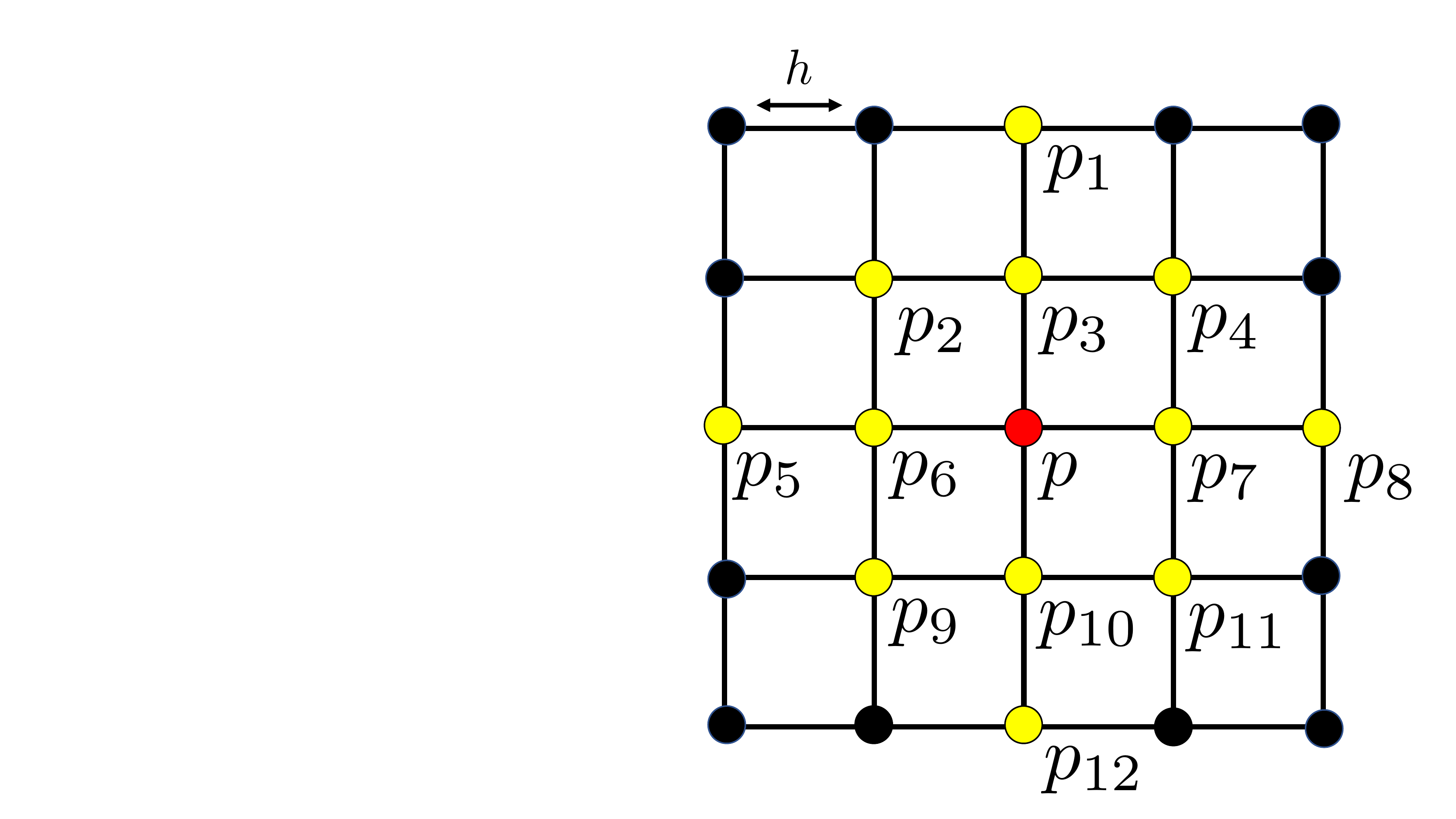} }}
\hfill
\subfloat[]{\def\scale1{.85}    \def\scale1{.85}
\begin{tikzpicture}[scale=\scale1, every node/.style={scale=\scale1}
,baseline=0pt
]
\coordinate (B1) at (1,.5);
\draw [thick] (B1) rectangle ($ (B1) + (1,3) $);
\node [rotate=90] at ($ (B1) + (.5,1.5) $) {Linear(128),ReLU};
\def\hx{.5}
\def\hy{3.25}
\draw [thick, ->] (\hx,\hy) -- (1,\hy);
\node [left, font=\large] at (\hx,\hy) {$h$};
\draw [thick, ->] (\hx,\hy-.5) -- (1,\hy-.5);
\node [left, font=\large] at (\hx,\hy-.5) {$u(p_1)$};

\node [left, font=\large] at (\hx,\hy-1) {$u(p_2)$};
\draw [thick, ->] (\hx,\hy-1) -- (1,\hy-1);
\node [left, font=\large] at (\hx,\hy-2.5) {$u(p_{12})$};
\draw [thick, ->] (\hx,\hy-2.5) -- (1,\hy-2.5);
\node [rotate=90, font=\huge] at (\hx,\hy-1.9) {$\bf \ldots$};
\coordinate (B2) at ($ (B1) + (1.5,0) $);
\draw [thick] (B2) rectangle ($ (B2) + (1,3) $);
\node [rotate=90] at ($ (B2) + (.5,1.5) $) {Linear(256),ReLU};
\draw [thick, ->] ($ (B1) + (1,1.5) $) -- ($ (B2) + (0,1.5) $);
\coordinate (B3) at ($ (B2) + (1.5,0) $);
\draw [thick] (B3) rectangle ($ (B3) + (1,3) $);
\node [rotate=90] at ($ (B3) + (.5,1.5) $) {Linear(128),ReLU};
\draw [thick, ->] ($ (B2) + (1,1.5) $) -- ($ (B3) + (0,1.5) $);
\coordinate (B4) at ($ (B3) + (1.5,0) $);
\draw [thick] (B4) rectangle ($ (B4) + (1,3) $);
\node [rotate=90] at ($ (B4) + (.5,1.5) $) {Linear(1),ReLU};
\draw [thick, ->] ($ (B3) + (1,1.5) $) -- ($ (B4) + (0,1.5) $);
\draw [thick, ->] ($ (B4) + (1,1.5) $) -- ($ (B4) + (1.5,1.5) $);
\node [right, font=\large] at ($ (B4) + (1.5,1.5) $) {$u(p)$};
\end{tikzpicture}}
\captionof{figure}{{\bf Deep Eikonal Solver for Cartesian Grids:}(a) For each point $p$, the network inputs the distances of its neighbors (colored yellow) along with the spacing $h$. (b) The network architecture with four fully connected layers that processes the local information from (a) and output the distance estimate $u(p)$.}
\label{fig:cartesian}
\vspace{-0.25cm}
\centering
{\includegraphics[width=0.94\textwidth]{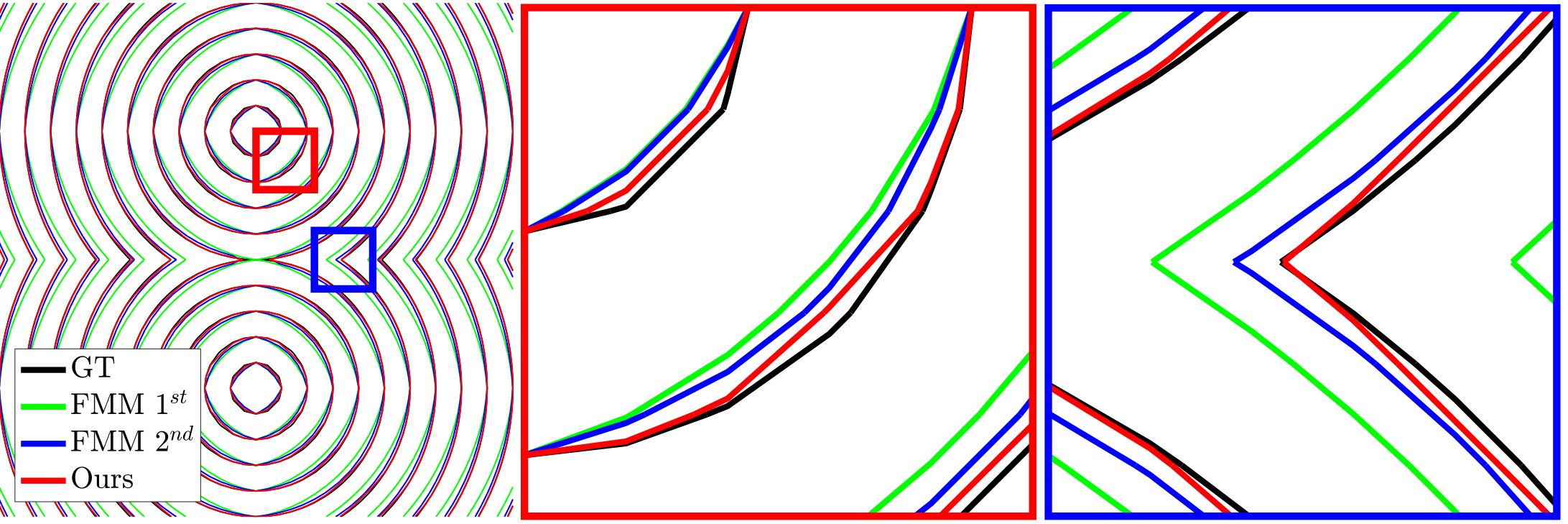}}
\vspace{0.1cm}
\captionof{figure}{{\bf Results on Euclidean grids:} Iso-contours of the Eikonal solution on 50x50 grid. The magnified plots show that the Deep Eikonal solver demonstrates better fidelity to the ground truth compared to the 1\textsuperscript{st} and 2\textsuperscript{nd} order fast marching schemes.}
\label{fig:result_cart}
\vspace{-0.4cm}
\end{table}
\subsection{Deep Eikonal Solver for Cartesian Grids}
\label{sect:cartesian}
We begin by evaluating our scheme on a Euclidean domain sampled regularly from $\Omega=[0,1]^2$. 
As the network's input patch for point $p: \mathcal{N}(p)$, we choose all the points located within a radius of $2h$ from $p$ (yellow points in Figure \ref{fig:cartesian}). 
We use a multilayer perceptron architecture (henceforth abbreviated as $mlp$) with four fully connected layers having number of nodes: $mlp(13,128,256,128,1)$ respectively. After each linear stage we applied ReLU function as the non-linearity. Since the grid spacing is uniform in the Cartesian case, $h$ encodes all necessary spatial information regarding neighborhood $\mathcal{N}(p)$. Accordingly, we input \emph{only} the grid spacing $h$ to the network instead of the explicit coordinates $\{p_1, p_2, \ldots,p_{12}\}$.
\begin{wrapfigure}{R}{0.4\textwidth}
\vspace{-0.4cm}
    \centering
    \includegraphics[width=0.4\textwidth]{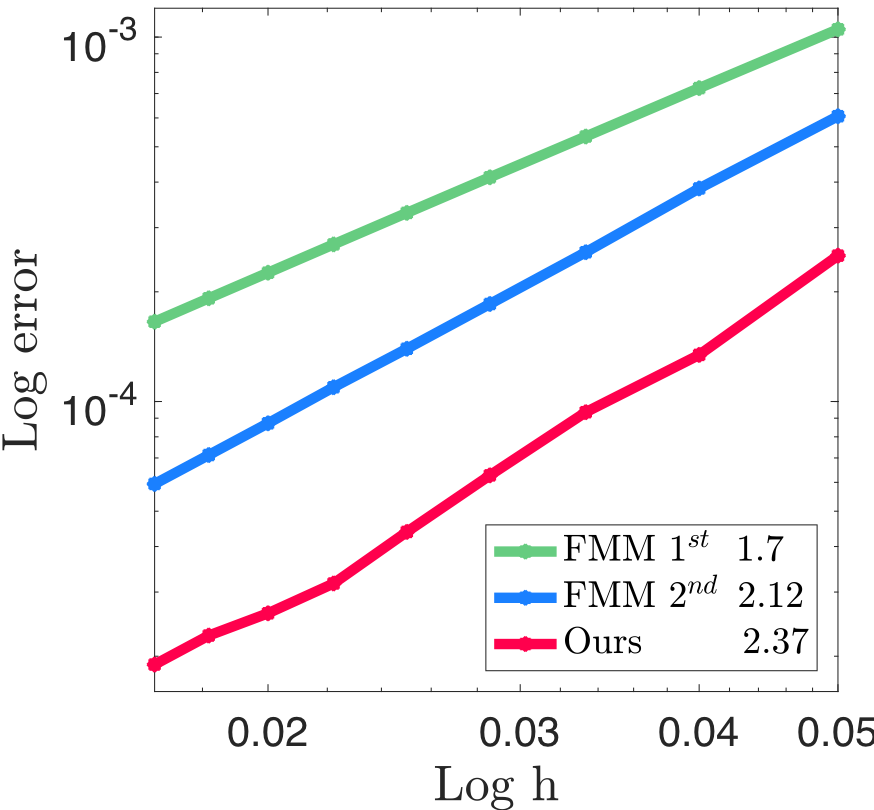}
\captionof{figure}{{\bf Order of Accuracy:} The slope of each error plot is the order of accuracy of the corresponding scheme.} \label{fig:grid_ooa}
\vspace{-0.75cm}
\end{wrapfigure}
We trained our network by generating 10,000 synthetic examples constructed as per the strategy enumerated in Section \ref{sect:train}. We design a variety of different source configurations like points, circles, arbitrary curves etc, and construct different local patches using the ground truth distance from these sources: $\mathfrak{s}$. For the Cartesian case, we evaluate the ground truth distance as
\begin{equation}
u_{gt}(p)= \min \{ \|p-s \| , s \in \mathfrak{s} \}.
\end{equation}
We pipeline each example by subtracting $\min \{ u_{gt}(q), q \in \mathcal{N}(p) \}$ from each distance and by scaling the example to mean magnitude 0.5, thereby simplifying the diversity of inputs prior to its feeding into the network.
Note, that the subtraction is equivalent to solving Equation (\ref{eq:eq1}) with corresponding initial condition $g'(x)=g(x) - \min \{ u_{gt}(q), q \in \mathcal{N}(p) \} $ and the scaling is equivalent to solving Equation (\ref{eq:eq1}) on correspondingly scaled coordinates.  
Before proceeding with the update step, the network's output is re-scaled and the bias is added to achieve the distance estimate.

We test our method using the benchmark shown in \cite{sethian1999level} comprising of two point sources. The learned scheme shows a superior and more accurate approximation of the distance function compared to fast marching local solvers (See Figure \ref{fig:result_cart}). 
We measure the order of accuracy $r$ of the proposed scheme by evaluating the error $\epsilon$ as a function of the grid spacing $h$ \cite{leveque1998finite}. Let $u_{gt}$ be the ground truth and $u_h$ represent the solution of a numerical scheme on a grid spacing $h$.   
\begin{eqnarray}
\epsilon(h) & = & |u_h - u_{gt}| \approx Ch^r + \mathcal{O}(h^{r+1})\\
\log(\epsilon) & \approx & \log(C) + r\log(h) + \mathcal{O}(h),
\end{eqnarray}
where $C$ is a constant. 
We evaluated our scheme for a range of grid sizes $h$ and plotted $\log\epsilon$ as a function of $\log h$. The slope of this line gives the order of accuracy $r$.
From Figure \ref{fig:grid_ooa} we can see that our local solver has a higher order of accuracy as compared to the classical fast marching local solvers in addition to lower errors.
\section{Deep Eikonal Solver for Triangulated Meshes}
\label{sect:mesh}
\begin{figure}[t]
    \centering
    \subfloat[]{\hspace{-1.3cm}{\includegraphics[width=6cm]{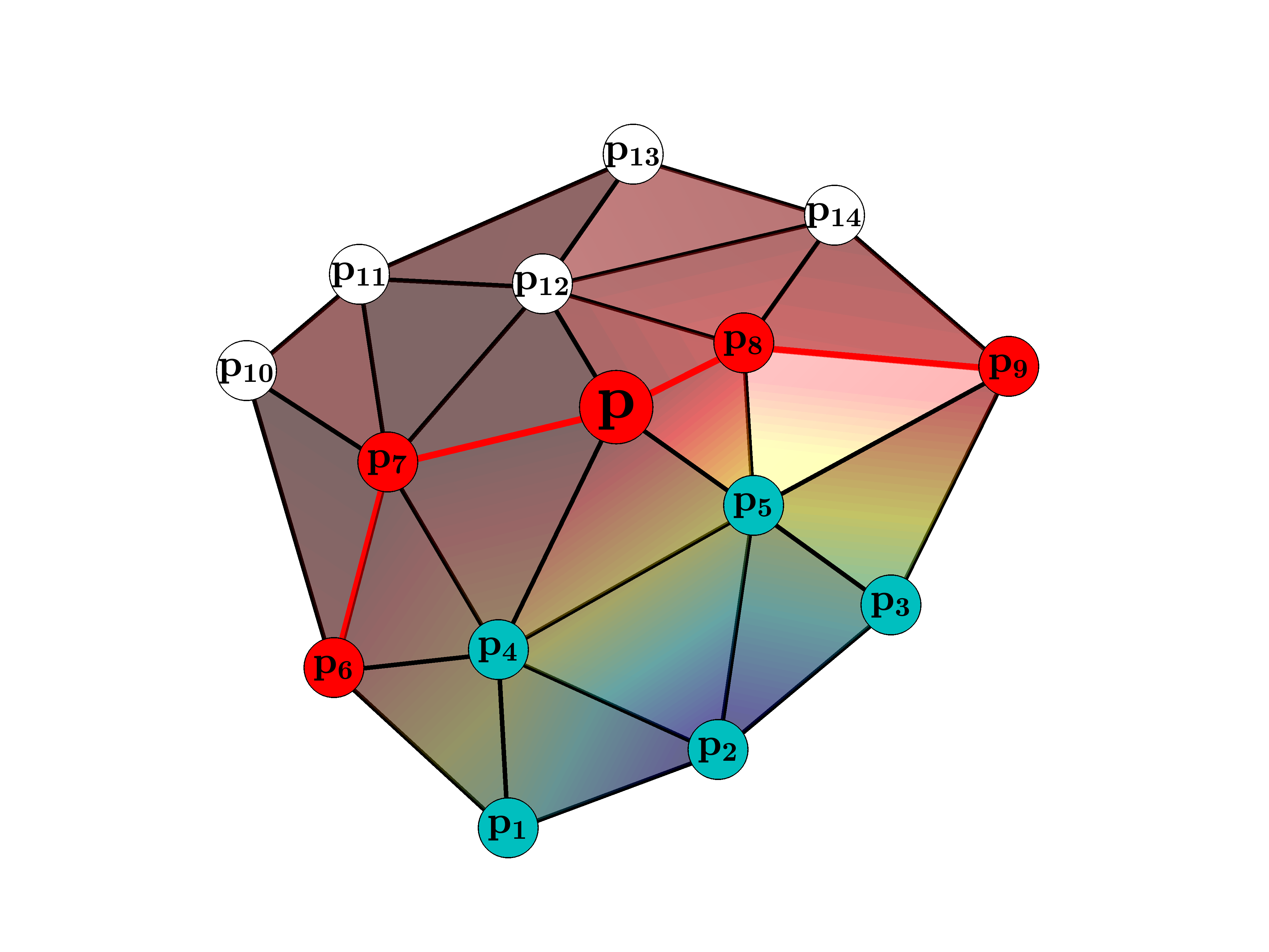} }}
    \qquad
    \def\scalemesh{.65}
    \subfloat[]{\hspace{-.8cm} \begin{tikzpicture}[scale=\scalemesh, every node/.style={scale=\scalemesh}
,baseline=-15pt
]
\def\hx{.5}
\def\hy{2.5}
\def\my{.5}
\draw [thick, ->] (\hx,\my) -- (1,\my);
\node [left, font=\large] at (\hx,\my) {$V_M$};
\coordinate (B1) at (1,0);
\draw [thick] (B1) rectangle ($ (B1) + (3,1) $);
\node [font=\large] at ($ (B1) + (1.5,0.5) $) {Linear, ReLU};
\node [rotate=90, font=\huge] at (\hx+2,\my+1.25) {$\bf \ldots$};
\def\y2{3}
\draw [thick, ->] (\hx,\y2) -- (1,\y2);
\node [left, font=\large] at (\hx,\y2) {$V_2$};
\coordinate (B3) at ($ (B1) + (0,\y2-.5) $);
\draw [thick] (B3) rectangle ($ (B3) + (3,1) $);
\node [font=\large] at ($ (B3) + (1.5,0.5) $) {Linear, ReLU};
\def\y1{4.5}
\draw [thick, ->] (\hx,\y1) -- (1,\y1);
\node [left, font=\large] at (\hx,\y1) {$V_1$};
\coordinate (B4) at ($ (B1) + (0,\y1-.5) $);
\draw [thick] (B4) rectangle ($ (B4) + (3,1) $);
\node [font=\large] at ($ (B4) + (1.5,0.5) $) {Linear, ReLU};
\draw [thick, ->] (4,.5) -- (4.5,.5);
\draw [thick, ->] (4,3) -- (4.5,3);
\draw [thick, ->] (4,4.5) -- (4.5,4.5);
\coordinate (B5) at (4.5,0);
\def\bw{1}
\def\bh{5}
\draw [thick] (B5) rectangle ($ (B5) + (\bw,\bh) $);
\node [rotate=90, font=\large] at ($ (B5) + (\bw/2,\bh/2) $) {Pooling};
\draw [thick, ->] ($ (B5) + (\bw,\bh/2) $) -- ($ (B5) + (\bw+0.5,\bh/2) $);
\coordinate (B6) at (6,\bh/2-1/2);
\def\bw{3}
\draw [thick] (B6) rectangle ($ (B6) + (\bw,1) $);
\node [font=\large] at ($ (B6) + (\bw/2,1/2) $) {Linear, ReLU};
\draw [thick, ->] (9,\bh/2) -- (9.5,\bh/2);
\node [right, font=\large] at (9.5,\bh/2) {$u(p)$};
\end{tikzpicture}
    }
    \caption{{\bf Deep Eikonal Solver for triangulated meshes:} (a) shows a local patch on a surface, the red line corresponds to the advancing ${\mbox{WaveFront}}$ passing from ${\mbox{Visited}}$ (green) to ${\mbox{Unvisited}}$ (white). (b) shows the network architecture (described in section \ref{sect:results}) for processing the local information. }
\label{fig:mesh}
\end{figure}

For triangulated meshes, we chose the second-ring neighbors of a point $p$ to constitute the local patch $\mathcal{N}(p)$ (See Figure \ref{fig:mesh}). 
In Euclidean domains, the regular sampling allows us to encode the spatial information of a patch using only the grid spacing $h$. 
However, the unordered nature of the sampled points within a mesh patch demands an explicit description of the spatial position associated with each point in a given patch. 
For each point $p_i$, we construct a vector comprising of four values: 
The 3D coordinates along with the distance $u(p_i)$,  namely, 
\begin{equation}
\label{eq:mesh_input}
V_i  = \big ( x(p_i)-x(p), y(p_i)-y(p), z(p_i)-z(p), u(p_i) \big ).
\end{equation}

The architecture we employ (see Figure \ref{fig:mesh}) has the following structure. 
It comprises of $M$ input vectors, each encoding the information about a single point in the neighborhood point of some patch as shown in Figure \ref{fig:mesh}. Each vector $V_{i}$ is processed independently in linear layers $mlp(4,64,128,512,1024)$ which produces a 1024-dimensional feature vector for each point. These 1024-dimensional feature vectors are max-pooled to generate a single vector of the same dimension and finally passed through a $mlp(512,256,1)$ which generates the desired output distance $u(p)$. Our architecture is motivated from previous deep learning frameworks for point clouds like \cite{qi2017pointnet}. This scheme of learning functions over non-Euclidean domains was analytically validated in \cite{zaheer2017deep}.
\subsection{Experimental results}
\label{sect:results}
We create $100,000$ training examples from 8 TOSCA shapes \cite{bronstein2008numerical} using the methodology described in Section \ref{sect:train}. In order to train for rotation invariance, we augment the data by multiplying each example by a random rotation matrix. As a pre-processing stage, we practice the same pipeline described in \ref{sect:cartesian}. 

\begin{figure}[tb]
\begin{center}
\makebox[\textwidth][c]{\includegraphics[width=0.8\textwidth]{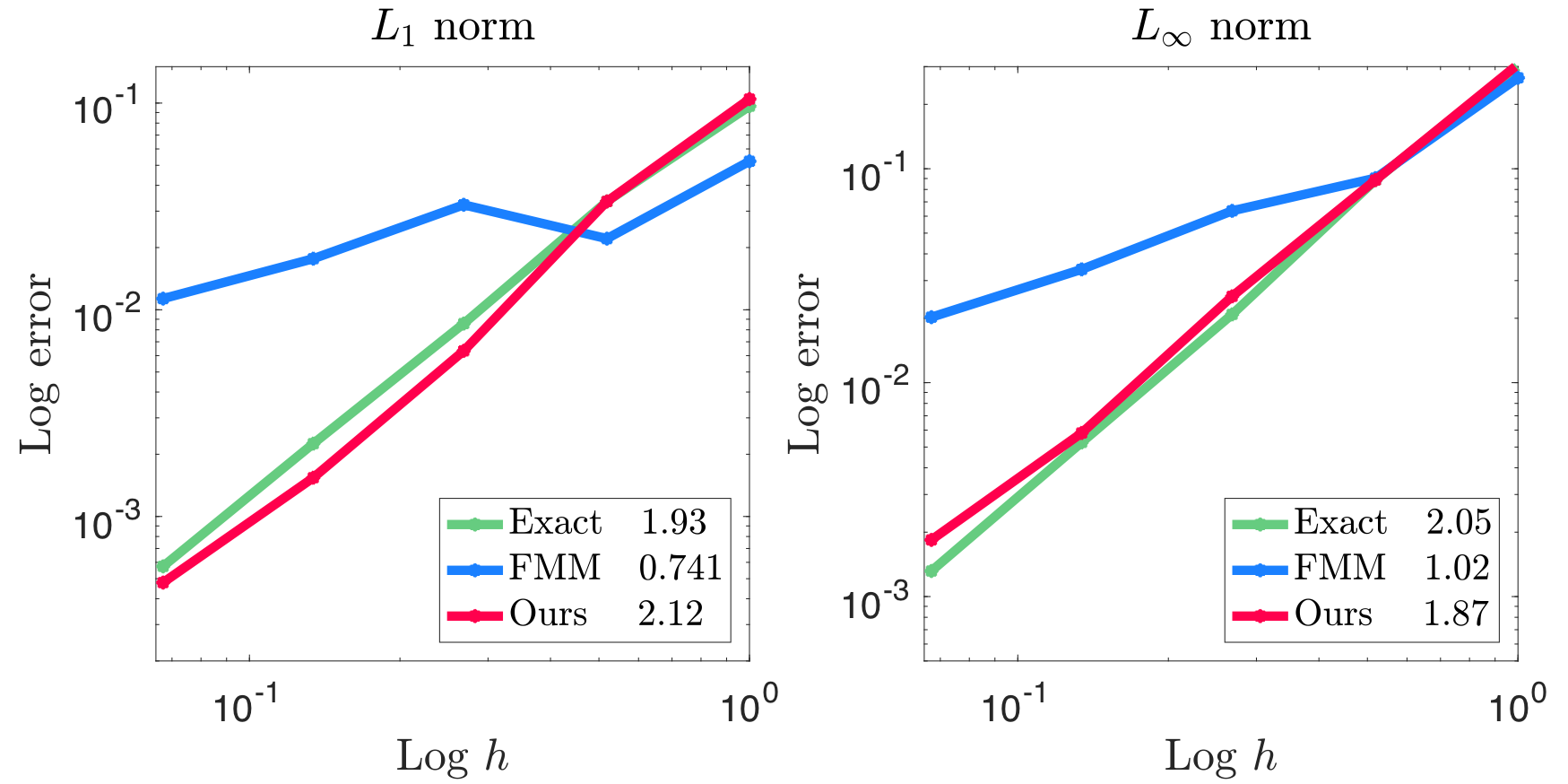}}
\caption{{\bf Order of Accuracy:} The plots shows the edge resolution impact on the error. Each scheme's accuracy associated with its corresponding slope. We construct a unit sphere mesh with various edge length using Loop's subdivision scheme.}
\label{fig:ooa}
\end{center}
\end{figure}

\begin{figure}[tb]
\centering
\includegraphics[width=0.95\textwidth]{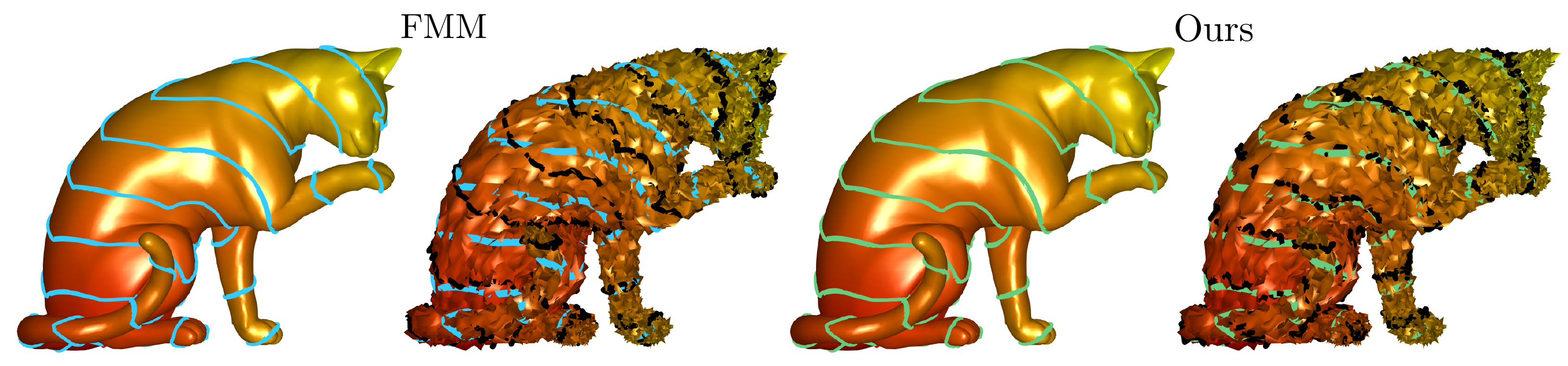}
\caption{{\bf Robustness to noise:} Equidistant isolines drawn on cat and its noisy version. The black isolines were computed on the noisy cat while the colorful isolines were computed on the original cat.}
\label{fig:noise}
\end{figure}

We compare our Deep Eikonal solver to two different axiomatic approaches: our direct competitor, the fast marching method and recent geodesic approximation, the heat method \cite{crane2013geodesics}. The heat method uses the heat equation solutions to estimate geodesics on surfaces, motivated by Varadhan's formula \cite{varadhan1967behavior}, that connects the two. As shown in Figures \ref{fig:vic} and \ref{fig:shrec}, the Deep Eikonal solver approximates the geodesic distance more accurately than the axiomatic approaches. Fast marching and heat method yield smooth but inaccurate approximations in comparison to the polyhedral scheme \cite{surazhsky2005fast} which is known to be the most accurate scheme for computing distances on surfaces.
Tables \ref{table:tosca} and \ref{table:shrec} show quantitative results on TOSCA \cite{bronstein2008numerical} and SHREC \cite{li2015comparison} databases respectively. The error at point $p$ was computed as $ | \frac{u(p) - u_{gt}(p)}{u_{gt}(p)} | $ where the ground truth is taken as the polyhedral scheme solution.
To evaluate our method's order of accuracy as described in Section \ref{sect:cartesian}, we use different resolutions of unit sphere, since the geodesic distance on a sphere can be computed analytically. Figure \ref{fig:ooa} demonstrates that the order of accuracy of the Deep Eikonal solver is comparable with the polyhedral scheme accuracy.
Finally, we test the robustness of our method to noise in the sampled manifold. Figure \ref{fig:noise} demonstrates that the isolines for our method remain robust to noise in comparison to the fast marching.

\begin{table}[p] 
\vspace{-0.25cm}
\centering
\includegraphics[width=10cm,height=4cm]{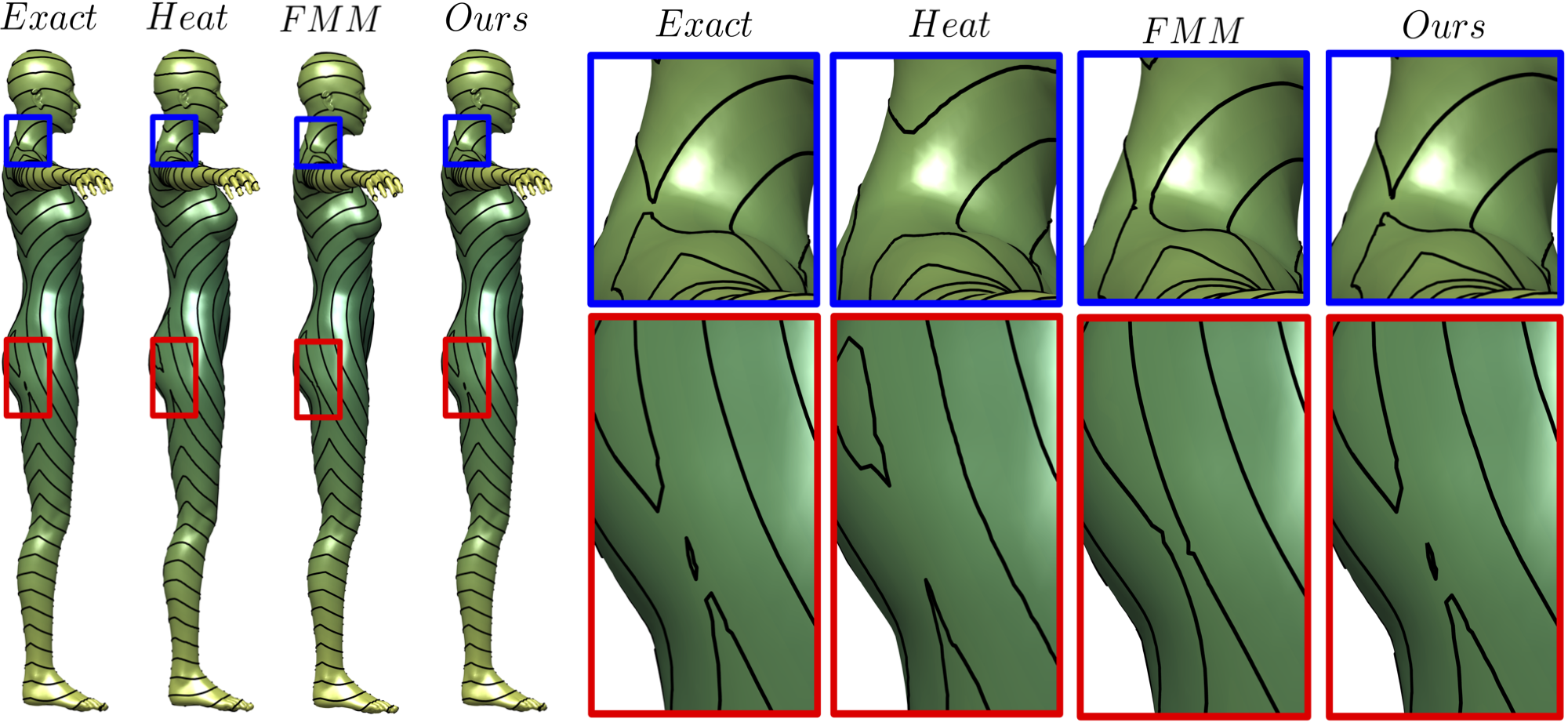}
\captionof{figure}{{\bf Intra-Dataset Generalization:} Iso-contours shown for (left to right) the polyhedral scheme, heat method, fast marching and our method. The network in our scheme was trained on patches extracted from 8 TOSCA shapes. The evaluation was conducted on a separate test set.}
\label{fig:vic}
\vspace{-0.2cm}
\newcolumntype{Y}{>{\centering\arraybackslash}X}
\begin{tabularx}{\textwidth}{ |Y|Y|Y|Y|Y|Y|Y| }
\cline{2-7}
\multicolumn{1}{c|}{} & \multicolumn{3}{|c|}{$L_1$} & \multicolumn{3}{|c|}{$L_{\infty}$} \\
\hline
\textbf{Model} & \textbf{Heat} & \textbf{FMM} & \textbf{Ours} & \textbf{Heat} & \textbf{FMM} & \textbf{Ours} \\
\hline 
\textbf{Cat}     & 0.2237 & 0.0060 & \textbf{0.0011} & 0.7775 & 0.0493 & \textbf{0.0352} \\
\hline
\textbf{Centaur} & 0.1185 & 0.0131 & \textbf{0.0042} & 0.3504 & 0.1622 & \textbf{0.1153} \\
\hline
\textbf{Dog}     & 0.0562 & 0.0121 & \textbf{0.0022} & 0.3393 & 0.2291 & \textbf{0.1034} \\
\hline
\textbf{Michael} & 0.0625 & 0.0082 & \textbf{0.0015} & 0.4196 & 0.2855 & \textbf{0.1042} \\
\hline
\textbf{Victoria}& 0.1129 & 0.0087 & \textbf{0.0015} & 0.6174 & 0.1885 & \textbf{0.0764} \\
\hline
\textbf{Wolf}    & 0.0254 & 0.0164 & \textbf{0.0054} & 0.2721 & 0.1465 & \textbf{0.0763} \\
\hline
\end{tabularx}
\vspace{0.1cm}
\captionof{table}{{\bf Intra-Dataset Generalization:} quantitative evaluation conducted on TOSCA. The errors were computed relative to the polyhedral scheme.}
\label{table:tosca}
\vspace{-0.25cm}
\includegraphics[width=12cm,height=4cm]{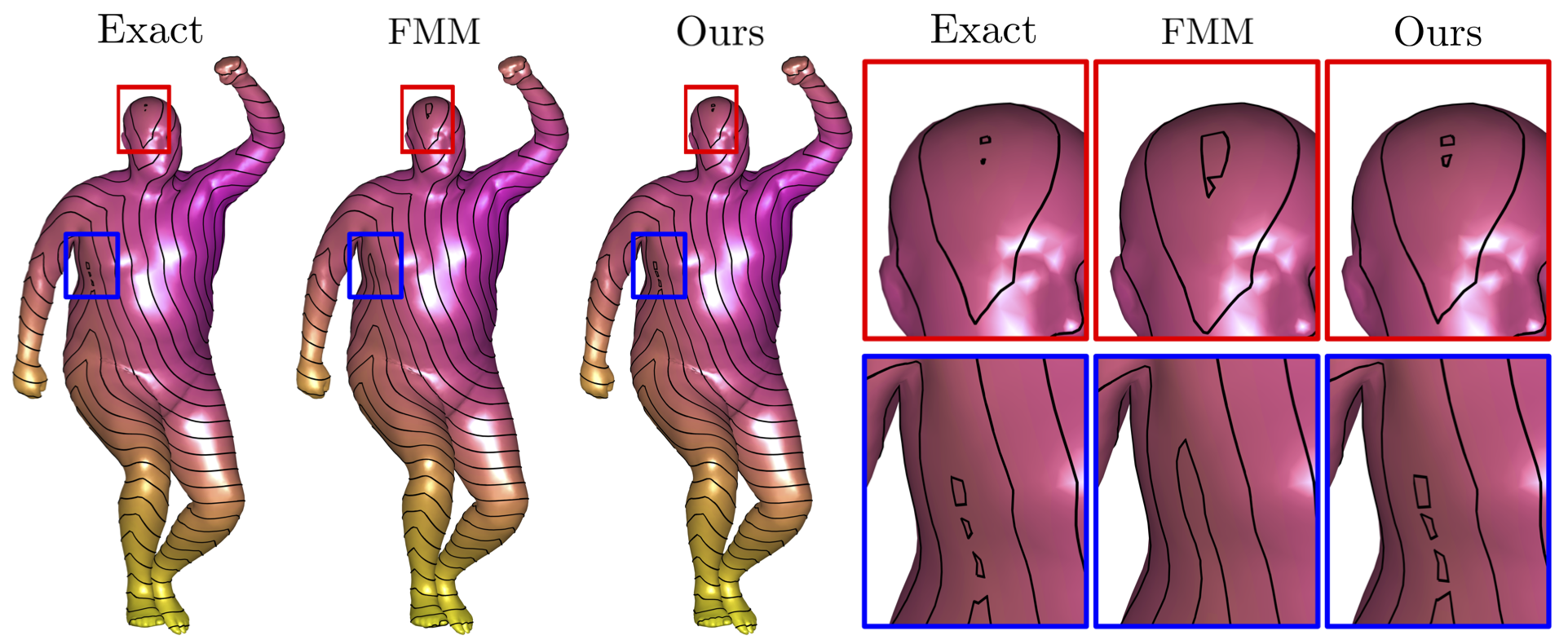}
\captionof{figure}{{\bf Inter-Dataset Generalization:} Iso-contours shown for (left to right) the polyhedral scheme, heat method, fast marching and our method. The evaluation was conducted on SHREC whereas the network was trained with TOSCA.}
\label{fig:shrec}
\vspace{-0.15cm}
\begin{tabularx}{\textwidth}{ |Y|Y|Y|Y|Y|Y|Y| }
\cline{2-7}
\multicolumn{1}{c|}{} & \multicolumn{3}{|c|}{$L_1$} & \multicolumn{3}{|c|}{$L_{\infty}$} \\
\hline
\textbf{Model} & \textbf{Heat} & \textbf{FMM} & \textbf{Ours} & \textbf{Heat} & \textbf{FMM} & \textbf{Ours} \\
\hline 
\textbf{Male 1}     & 0.0288 & 0.0105 & \textbf{0.0027} & 0.2570 & 0.0981 & \textbf{0.0681} \\
\hline
\textbf{Male 2}     & 0.0152 & 0.0122 & \textbf{0.0026} & 0.1553 & 0.1694 & \textbf{0.0296} \\
\hline
\textbf{Male 3}     & 0.0274 & 0.0105 & \textbf{0.0048} & 0.1926 & 0.1479 & \textbf{0.0762} \\
\hline
\textbf{Female 1}   & 0.0236 & 0.0089 & \textbf{0.0034} & 0.3184 & 0.1535 & \textbf{0.0504} \\
\hline
\textbf{Female 2}   & 0.0229 & 0.0083 & \textbf{0.0019} & 0.1350 & 0.0970 & \textbf{0.0383} \\
\hline
\textbf{Female 3}   & 0.0410 & 0.0087 & \textbf{0.0026} & 0.4205 & 0.2070 & \textbf{0.0671} \\
\hline
\end{tabularx}
\vspace{0.1cm}
\captionof{table}{{\bf Inter-Dataset Generalization:} quantitative evaluation conducted on SHREC. The errors were computed relative to the polyhedral scheme.}\label{table:shrec}
\end{table}

\section{Conclusion}
\vspace{-0.2cm}
In this paper, we introduce a new methodology by which we train a neural network to numerically solve the Eikonal equation on regularly and non-regularly sampled domains. 
We develop a numerical scheme which uses a data-centric learning-based computation in conjunction with a well-established axiomatic update method. 
In comparison to the axiomatic counterparts, we demonstrate that our hybrid scheme results in a considerable gain in performance measured by lower errors and larger orders of accuracy tested on a variety of different settings. 
Our approach advocates combining the best of both worlds: approximation power of neural networks combined by meaningful axiomatic numerical computations in order to achieve performance as well as better understanding of computational algorithms. 
\vspace{-0.2cm}



\end{document}